\documentclass{article}

\usepackage{microtype}
\usepackage{graphicx}
\usepackage{subfigure}
\usepackage{booktabs} 
\usepackage{multirow}
\usepackage[table]{xcolor}  
\usepackage{pifont}
\definecolor{mred}{RGB}{238, 34, 12}
\definecolor{mgreen}{RGB}{1, 127, 0}
\definecolor{mblue}{RGB}{0, 77, 158}
\newcommand{\mredbf}[1]{\textcolor{mred}{\textbf{#1}}}
\newcommand{\mbluebf}[1]{\textcolor{mblue}{\textbf{#1}}}

\usepackage{amsmath}
\usepackage{arydshln} 
\usepackage{caption}   
\usepackage{hyperref}

\usepackage[accepted]{icml2025}

\usepackage{amsmath}
\usepackage{amssymb}
\usepackage{mathtools}
\usepackage{amsthm}
\usepackage{titlesec}
\usepackage[capitalize,noabbrev]{cleveref}

\theoremstyle{plain}

\theoremstyle{definition}

\theoremstyle{remark}

\usepackage[textsize=tiny]{todonotes}

\titlespacing*{\section}{0pt}{1pt plus 0.2pt minus 0.5pt}{1pt plus 0.2pt minus 0.5pt} 
\titlespacing*{\subsection}{0pt}{1pt plus 0.2pt minus 0.5pt}{1pt plus 0.2pt minus 0.5pt} 
\titlespacing*{\paragraph}
{0pt}    
{0.5em}    
{0.5em}    
\begin{document}

\twocolumn[
\icmltitle{MagicWand: A Universal Agent for Generation and Evaluation Aligned with User Preference}


\icmlsetsymbol{equal}{*}

\begin{icmlauthorlist}
\icmlauthor{Zitong Xu}{equal,yyy}
\icmlauthor{Dake Shen}{equal,comp}
\icmlauthor{Yaosong Du}{comp}
\icmlauthor{Kexiang Hao}{comp}
\icmlauthor{Jinghan Huang}{sch}
\icmlauthor{Xiande Huang$^{\dagger}$}{comp} \\
\icmlauthor{\textsuperscript{1}Shanghai Jiao Tong University, Shanghai, China}{}
\icmlauthor{\textsuperscript{2}De Artificial Intelligence Lab, Hainan, China}{}\\
\icmlauthor{\textsuperscript{3}The Chinese University of Hong Kong, Shenzhen China}{}\\
\icmlauthor{\textsuperscript{*}Equal Contribution}{} \icmlauthor{\textsuperscript{$^{\dagger}$}Corresponding Author}{}

\end{icmlauthorlist}

\icmlaffiliation{yyy}{Shanghai Jiao Tong University, Shanghai, China}
\icmlaffiliation{comp}{De Artificial Intelligence Lab, Hainan, China}

\icmlcorrespondingauthor{Xiande Huang}{xdhuang@dail.email}

\icmlkeywords{Machine Learning, ICML}
\begin{center}
    \centering
    \includegraphics[width=1\linewidth]{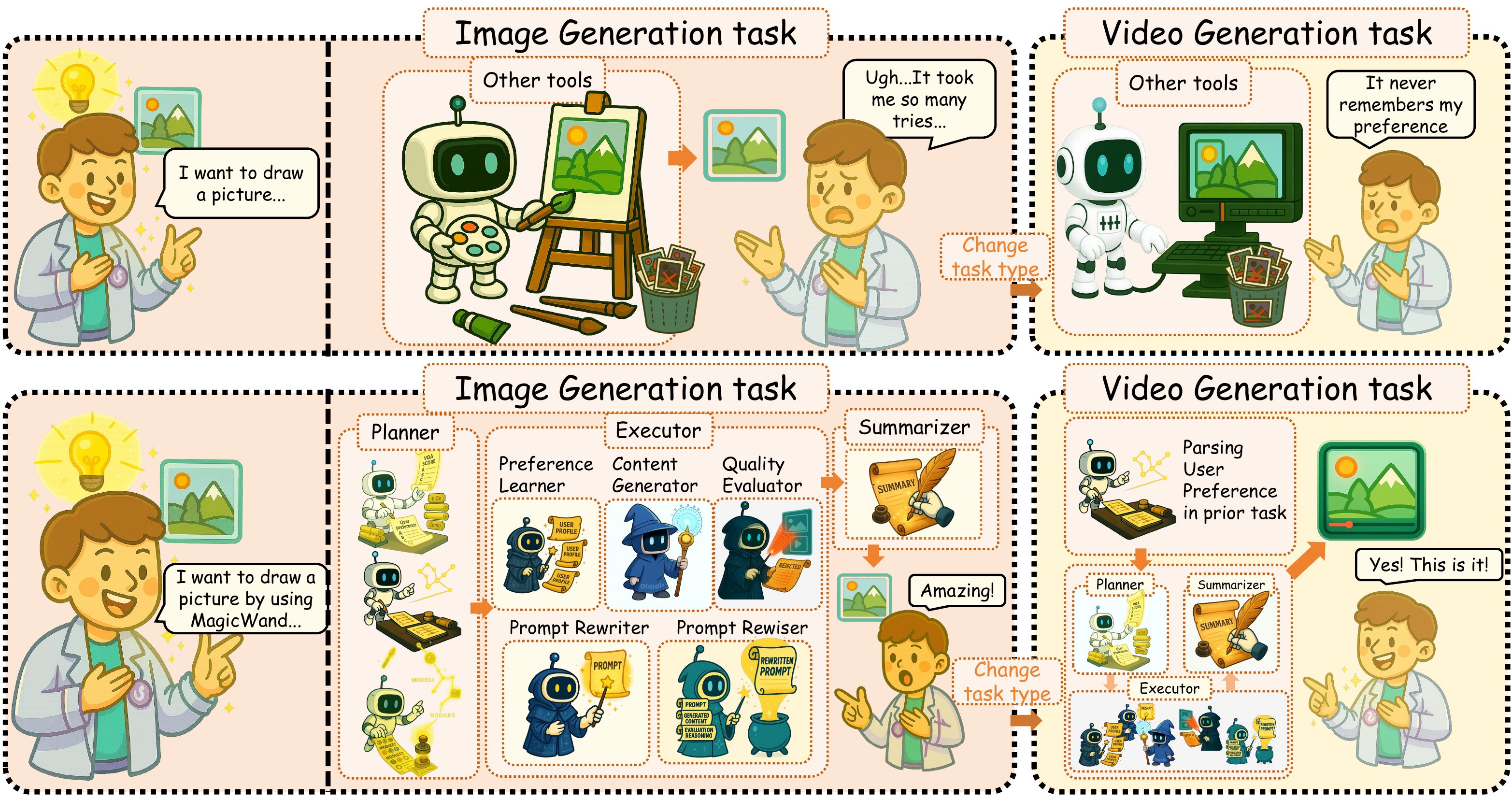}
\captionof{figure}{Comparison of existing AIGC methods and our MagicWand in application.}
    \label{problem}
\end{center}
\vskip 0.3in
]




\begin{abstract}
Recent advances in AIGC (Artificial Intelligence Generated Content) models have enabled significant progress in image and video generation. However, users still struggle to obtain content that aligns with their preferences due to the difficulty of crafting detailed prompts and the lack of mechanisms to retain their preferences. To address these challenges, we construct \textbf{UniPrefer-100K}, a large-scale dataset comprising images, videos, and associated text that describes the styles users tend to prefer. Based on UniPrefer-100K, we propose \textbf{MagicWand}, a universal generation and evaluation agent that enhances prompts based on user preferences, leverages advanced generation models for high-quality content, and applies preference-aligned evaluation and refinement. In addition, we introduce \textbf{UniPreferBench}, the first large-scale benchmark with over 120K annotations for assessing user preference alignment across diverse AIGC tasks. Experiments on UniPreferBench demonstrate that MagicWand consistently generates content and evaluations that are well aligned with user preferences across a wide range of scenarios.
\end{abstract}
\section{Introduction}
The rapid development of AIGC (Artificial Intelligence Generated Content) models has significantly advanced the generation of high-quality image and video content. For example, Qwen-Image \cite{qwenedit} can generate text-aligned images, NanoBanana \cite{nanolink} can perform realistic image editing, Sora2 \cite{sora} excels in video generation, and Gen4 supports video editing. Nevertheless, users still often encounter challenges in obtaining outputs that meet their intended preferences \cite{enhancer}. This limitation is not due to the generative capabilities of these models, but can be attributed to three fundamental issues of current AIGC systems: \textbf{(i) Limited expressivity of user prompts:} as non-expert users often struggle to precisely articulate content requirements, resulting in under-specified or ambiguous prompts that lead to outputs diverging from the user’s intended style or content \cite{enhancer}. \textbf{(ii) Absence of user preference modeling:} since existing generation methods lack mechanisms to retain or adapt to individual preferences across interactions, resulting in inconsistent outputs and limited alignment with evolving user requirements. \textbf{(iii) Fragmented support across generation tasks:} as current approaches typically address isolated generation modalities rather than providing a unified framework, preventing users from achieving coherent, preference-aligned results across diverse content types.

As AIGC models continue to advance, numerous studies have focused on evaluation and benchmarking, often providing comprehensive coverage over large-scale datasets \cite{love, vbench,vebench,lmm4edit,lmm4lmm,tdve}. Some of these works further investigate alignment with human perception. Nevertheless, as shown in Figure~\ref{problem}, they still face several critical limitations: \textbf{(i) Lack of personal preference evaluation:} Existing evaluations mainly capture aggregate human preference using metrics like mean opinion scores (MOS) \cite{lmm4lmm,love,lmm4edit}, which fail to reflect individual user tastes. \textbf{(ii) Limited cross-task evaluation:} Most benchmarks focus on a single generation modality, restricting the assessment of model performance across diverse AIGC tasks; and \textbf{(iii) Limited interpretability:} Evaluation results provide little insight into why certain outputs are preferred, hindering personalized content optimization.

To address these limitations, we first construct \textbf{UniPrefer-100K}, which contains 100K AI-generated images and videos, each annotated with text describing the preferred style (\textit{e.g.,} bright and vivid color palettes, realistic lighting, cinematic composition, and a cozy or dreamy atmosphere). Based on UniPrefer-100K, we propose \textbf{MagicWand}, the first agent for universal generation and evaluation. It can produce and assess content aligned with user preferences across diverse AIGC tasks, including image generation, image editing, text-guided video generation, image-guided video generation, and video editing. MagicWand follows a modular design consisting of three main components: Planner, Executor, and Summarizer.
    





To validate the effectiveness of our MagicWand, we have introduced \textbf{UniPreferBench}, the first large-scale benchmark for evaluating user preference alignment across multiple AIGC tasks. UniPreferBench contains 5 types of generation tasks, 24 diverse generation models, and 120K user preference annotations. It provides a standardized and comprehensive platform for assessing preference-aware performance across AIGC systems and its evaluation methods. Experimental results based on UniPreferBench demonstrate that MagicWand consistently generates content aligned with user preferences while providing human-aligned, interpretable evaluations across different generation scenarios. In summary,our main contributions are:
\begin{itemize}
    \item We construct UniPrefer-100K, a large-scale dataset comprising 100K AI-generated contents accompanied by textual preference descriptions.
    \item We propose MagicWand, the first agent for multi-task generation and evaluation, capable of producing outputs aligned with user preferences along with interpretable feedback.
    \item We introduce UniPreferBench, the first large-scale benchmark containing 120K user preference scores across different AIGC tasks, enabling standardized cross-task preference evaluation.
    \item Extensive experiments demonstrate that MagicWand consistently outperforms state-of-the-art methods for both content generation and evaluation in preference alignment.
\end{itemize}
\begin{figure*}[t!]
\centering
  \includegraphics[width=1\textwidth]{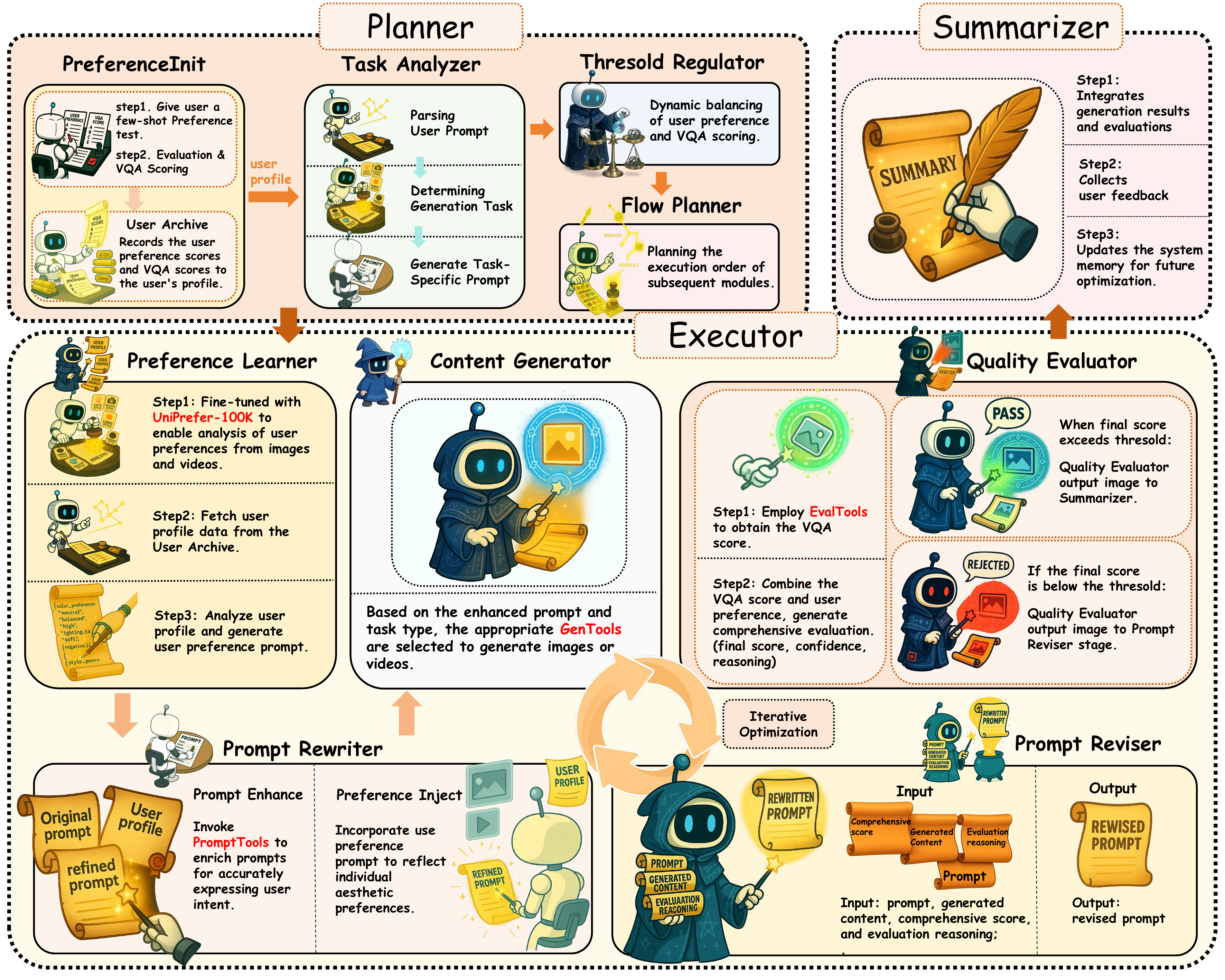}
  \caption{Overview of our MagicWand.}
  \label{agent}
\end{figure*}
\section{Related Work}
\subsection{Artificial Intelligence Generated Content}
\textbf{Image generation (T2I).}
Image generation refers to the process of creating images from textual descriptions \cite{t2isurvey}. This field has rapidly evolved from early GAN-based approaches to the more recent diffusion-based models \cite{t2isurvey}, which have set new standards for realism and diversity. Diffusion models iteratively refine random noise into coherent visual structures and support flexible conditioning \cite{SD,sdxl}. By combining diffusion with transformer backbones, modern generative models not only achieve higher visual fidelity but also enhance semantic understanding \cite{ditt}, enabling fine-grained control.

\textbf{Image editing (I2I).}
Image editing methods aim to modify a source image based on editing instructions while preserving realism and consistency with surrounding regions \cite{ESurvey}. With the advent of diffusion-based architectures, editing has become increasingly precise and interpretable, supporting localized operations such as object addition, replacement, and color transfer \cite{nanobanana,fluxkontext,seedream4}. The incorporation of transformer structures into diffusion editors further enhances contextual reasoning, enabling models to capture spatial and semantic relationships within an image \cite{qwenedit,omnigen2}. This combination allows for controllable, high-fidelity edits that closely align with user intent.

\textbf{Text-guided video generation (T2V).}
Text-to-video generation extends the success of text-to-image models to the temporal domain \cite{t2vsurvey}. Diffusion-based video generators adapt spatial denoising to spatiotemporal processes, ensuring both frame quality and temporal coherence \cite{love,cogvideox}. By leveraging transformer-like temporal attention modules, these systems can maintain consistent motion, object appearance, and scene dynamics across frames. This combination of diffusion and transformer reasoning allows natural language prompts to drive complex visual narratives with coherent motion and structure \cite{hunyuanvideo}.

\textbf{Image-guided video generation (I2V).}
Image-guided video generation aims to synthesize videos from a given reference image, ensuring that the generated video maintains both appearance consistency and plausible motion dynamics \cite{vbench}. This process involves generating temporally coherent frames that preserve the visual style, composition, and features of the reference image while introducing natural, smooth transitions across the video sequence \cite{hunyuanvideo}. Recent advancements in generative models have leveraged the combination of diffusion and transformer architectures to tackle this challenge, enabling models to capture not only the static visual characteristics but also the dynamic temporal evolution of the scene \cite{cogvideox,sora}. By coupling image features with learned temporal priors, these hybrid models effectively integrate spatial and temporal information, allowing them to generate realistic, high-fidelity videos that remain visually consistent and stable over time.

\textbf{Video editing (V2V).}
Video editing refers to the process of modifying and enhancing temporal sequences of videos, enabling flexible and precise alterations based on user-defined instructions. Current video editing approaches can be broadly categorized into two types. Inversion-based methods avoid the need for paired video-text-edit data, instead leveraging inversion techniques to directly manipulate video content \cite{tune,fatezero}. On the other hand, end-to-end feed-forward models rely on large-scale human-annotated video edit data to generate high-quality video modifications. These models are capable of handling more complex edits and maintaining higher visual fidelity \cite{insvie, insv2v, ditt}.

However, existing AIGC methods are typically specialized for a one or two generation task and lack a unified framework capable of integrating these diverse generation tasks. This fragmentation limits the ability to perform cross-task reasoning and consistent preference alignment.

\subsection{Visual Quality Assessment}
Traditional visual quality assessment (VQA) methods were primarily designed to evaluate the perceptual quality of images degraded by distortions such as blur, noise, compression artifacts, or transmission errors \cite{DBCNN, Hyper, MANIQA,TOPIQ}. These methods have proven effective for assessing natural distortions but often fail to capture higher-level semantic consistency in content generated by modern generative models. With the rapid proliferation of AIGC, the focus of VQA has expanded beyond distortion measurement to the evaluation of generated content \cite{lmm4lmm, tdve, love, lmm4edit}, considering semantic alignment, structural coherence, detail realism, and overall authenticity. These studies have also explored human preference VQA based on subjective mean opinion scores (MOSs), but they generally focus on a single type of AIGC task and emphasize aggregate human preferences rather than personalized aesthetic tendencies. Consequently, current VQA research lacks mechanisms to model individual-level preference diversity and to evaluate visual quality consistently across different generative modalities.
\subsection{Agent}
The rapid progress of large language models (LLMs) has endowed them with advanced reasoning, planning, and interaction capabilities, enabling their integration as central controllers of autonomous agents \cite{llmagent1,llmagent2}. These agents are not only capable of perceiving environments and executing complex tasks but also of maintaining long-term memory, adapting to user feedback, and continuously improving their behavior through self-evolution \cite{agent3}. Recent developments in agentic systems further highlight their ability to decompose tasks, collaborate across roles, and refine outputs through iterative reflection, making them particularly effective for complex multimodal scenarios \cite{agent1, agent2}. Building on these strengths, we introduce an agentic framework that leverages dynamic planning, memory-driven adaptation, and multimodal interaction to achieve personalized, interpretable, and preference-aligned generation and evaluation across AIGC tasks.

\section{UniPrefer-100K}
UniPrefer-100K spans five major AIGC tasks, including T2I, I2I, T2V, I2V, and V2V, containing 25K samples for each task. We first curate an initial pool of 50K candidate samples per task. For T2I, images are sourced from EvalMi-50K \cite{lmm4lmm}. For I2I, images are drawn from Pico-Banana-400K \cite{apple}. For T2V, videos are obtained from AIGVE-60K \cite{love}. For I2V, videos are collected from VBench \cite{vbench} and further augmented using HunYuanVideo-I2V \cite{hunyuanvideo}. For V2V, videos originate from TDVE-DB \cite{tdve} and are supplemented with additional samples generated by Ditto \cite{ditt}.

To generate preference text, we first employ Qwen3-VL \cite{qwen3} to infer user-preferred styles based on attributes such as lighting, color, composition, and overall aesthetic cues. We then recruit 50 human annotators to review and refine the model-generated descriptions. During this process, low-quality or unsuitable samples are discarded, resulting in approximately half of the initial pool being filtered out. The remaining samples are manually corrected and annotated with high-quality preference text that accurately captures user-preferred stylistic characteristics. Finally, we curate 100K images and videos with associated preference text, providing a large-scale resource that connects AIGC outputs with user-preference textual descriptions.

\section{MagicWand}
As shown in Figure~\ref{agent}. The MagicWand architecture is composed of three principal components, each leveraging a multimodal large language model (MLLM): the Planner, Executor, and Summarizer. These components operate within a structured reasoning workflow, where each module performs a specific function by exploiting the capabilities of MLLM.
\subsection{Planner: Task and Preference Management}
The reasoning process begins with the construction of a task- and preference-aware generation plan. Given a user prompt $u$, optional input content $x$ (image or video), the Planner module $P$ acts as a task interpreter and workflow manager, producing a structured plan $P_u(x)$ that guides subsequent generation and evaluation steps. The planning procedure consists of four components:

\textbf{(1) Task Analyzer:}
The Task Analyzer parses the user prompt $u$ to determine the intended generation type $\tau \in \mathcal{T} = \{\text{T2I}, \text{I2I}, \text{T2V}, \text{I2V}, \text{V2V}\}$
and extracts the task-specific generation prompt $t$ for downstream modules. 

\textbf{(2) Preference Manager:}
To initialize user preferences, MagicWand employs few-shot sampling, typically collecting five samples per generation task. For each sample, the user provides a score, which is combined with system-generated evaluations from the integrated VQA tools. These data are stored in memory to establish an initial user preference representation that guides subsequent adaptive decision-making. Furthermore, after each generation, the user’s score of the generated result and the corresponding VQA evaluation are also recorded and added to memory, continuously refining the user preference representation over time. The generated samples $y^\tau$ in type $\tau$ and its preference score $p^\tau$ and VQA score $v^\tau$ together stored in memory $M$.

\textbf{(3) Threshold Regulator:}
The Threshold Regulator dynamically sets the evaluation threshold for generation type $\tau$, combining user preference scores, VQA scores. The threshold is computed as:
\begin{equation}
\begin{aligned}
Threshold^\tau = 
& \; \beta_1 \cdot \frac{1}{N_\tau} \sum_{i=1}^{N_\tau} \big(v_i^\tau - p_i^\tau\big) \\
& + \beta_2 \cdot \sum_{\substack{\tau' \in \mathcal{T} \\ \tau' \neq \tau}} \frac{1}{N_{\tau'}} \sum_{i=1}^{N_{\tau'}} \big(v_i^{\tau'} - p_i^{\tau'}\big) \\
& + \frac{1}{N} \sum_{i=1}^{N} p_i
\end{aligned}
\end{equation}
where $\beta_1$ and $\beta_2$ are preference coefficients that control the influence of user preferences on the evaluation threshold. Specifically, $\beta_1$ represents the intra-task preference coefficient, governing the impact of user feedback within the current generation task $\tau$, while $\beta_2$ represents the cross-task preference coefficient, capturing the influence of user preferences from other generation tasks to adjust the threshold adaptively. When no user preference is available, the threshold reduces to the average VQA score across historical samples, providing a neutral baseline for evaluation. Once user feedback is incorporated, the threshold adapts dynamically: if the user score $v_i$ is lower than the corresponding VQA score $p_i$, the threshold increases, effectively raising the standard to better satisfy the user’s expectations; conversely, if the user score is higher than the VQA score, the threshold decreases, relaxing the standard to allow faster generation. This adaptive mechanism ensures that the system balances objective quality assessment from the VQA tools with subjective user satisfaction, enabling personalized and efficient content refinement.

\textbf{(4) Flow Planner:}
The Flow Planner is responsible for orchestrating the execution sequence of the MagicWand modules, ensuring that each module operates in a coherent and efficient order. It coordinates the interactions between modules, guiding iterative refinement when generated content does not meet the evaluation threshold. By managing this structured workflow, the Flow Planner enables the system to adaptively optimize content quality while maintaining efficient processing across different generation tasks.
\subsection{Executor: Sub-task Execution and Tool Invocation}
Given the Plan $P_u(x)$, generation type $\tau$, and user preference memory $M$, the Executor module $E$ sequentially performs the sub-tasks, generating the contents 
\begin{equation}
    C=E(x, t, \tau, P_u(x), M)
\end{equation}
It serves as the central operational module of MagicWand, completing the workflow from prompt refinement to content generation, to quality evaluation and adaptive revision. It sequentially invokes the Preference Learner, Prompt Rewriter, Content Generator, Quality Evaluator, and Prompt Reviser, enabling a closed-loop generation process. Through iterative reasoning and evaluation, the Executor ensures that each generated result is progressively optimized to align with user preferences, maintain visual quality.

\textbf{(1) Preference Learner:}
The Preference Learner extracts user preference representations from the memory $M$. We have fine-tuned the MLLM, Qwen3-VL \cite{qwen3},  using our UniPrefer-100K dataset, enabling it to analyze AI-generated images and videos and generate corresponding user preference descriptions. Given samples in $M$, the model infers a preference prompt that captures the user’s generation tendencies (\textit{e.g.,} favoring bright and vivid image styles). This prompt serves as a semantic bridge between user intent and controllable generation parameters, guiding subsequent modules toward personalized outputs. This process can be expressed as 
\begin{equation}
    t_p = PreferenceLearner(M)
\end{equation}
where $t_p$ is the user preference prompt.

\textbf{(2) Prompt Rewriter:}
The Prompt Rewriter refines the original generation prompt $t$ by integrating the inferred preference prompt $t_p$ and prompt tools. Specifically, the PromptTool includes PromptEnhancer \cite{enhancer} for T2I, PromptEnhancer-img2img-edit \cite{enhancer} for I2I, Qwen3-VL \cite{qwen3} for other generation types. With the generation type $\tau$, the selected prompt tool is $PromptTools(\tau)$. The extended prompt $t_e$ can be expressed as
\begin{equation}
    t_e = PromptRewriter(t_p, PromptTools(\tau))
\end{equation}
This module can produce a contextually enhanced prompt that better reflects the user’s implicit aesthetic or semantic intent, solving the problem that non-professional users are hard to express the intention. This step improves controllability and coherence, ensuring that the generated content aligns more closely with user preference.

\textbf{(3) Content Generator:}
The Content Generator executes the actual generation process. Given a generation type $\tau$, it invokes corresponding state-of-the-art generative tools to produce candidate results. Specifically, for image generation (T2I), it uses the open-source Qwen-Image \cite{qwenedit} and the closed-source SeedDream4 \cite{seedream4}. For image editing (I2I), the module utilizes open-source Qwen-Edit \cite{qwenedit} and closed-source NanoBanana \cite{nanobanana}. For text-guided video generation (T2V), open-source HunYuanVideo \cite{hunyuanvideo} and closed-source Sora are employed. In the case of image-guided video generation (I2V), open-source HunYuan-I2V \cite{hunyuanvideo} and closed-source Sora2 \cite{sora} are used. For video editing (V2V), open-source Ditto \cite{ditt} and closed-source Gen4 \cite{gen4} are selected. For each generation type $\tau$, the user can select either an open- or closed-source generation model, denoted as $GenTools(\tau, s)$, where $s \in \{\text{Open}, \text{Closed}\}$. By integrating both open-source and proprietary tools across modalities, the Content Generator ensures flexibility and allows systematic benchmarking across state-of-the-art models. It takes as input either extended prompt $t_e$ from Prompt Rewriter or revised prompt $t_r$ from Prompt Reviser, the generation type $\tau$, generating the image or video 
\begin{equation}
    y^\tau = ContentGenerator(t_{e/r}, GenTools(\tau,s))
\end{equation}

\textbf{(4) Quality Evaluator:}
The Quality Evaluator combines state-of-the-art quality assessment methods from integrated EvalTools with user preference information to produce a comprehensive evaluation score and an interpretable reasoning statement. Specifically, EvalTools utilizes LMM4LMM \cite{lmm4lmm} for T2I, LMM4Edit \cite{lmm4edit} for I2I, LOVE \cite{love} for T2V, VBench-I2V \cite{vbench} for T2V and TDVE-assessor \cite{tdve} for V2V. Given an source image/video $x$, generation prompt $t$, generated image/video $y^\tau$, the VQA score $v^\tau$ is generated by $v^\tau = EvalTools(x,t,y^\tau)$. Quality Evaluator then takes $y^\tau$, $v^\tau$ and user preference prompt $t_p$ as inputs, generate the final evaluation score $f^\tau$ and interpretable evaluation reasoning $r$ using another MLLM. The entire process can be expressed as 
\begin{equation}
    f^\tau, r = QualityEvaluator(y^\tau, t_p, EvalTools(x,t,y^\tau))
\end{equation}
If the final evaluation score $f^\tau$ falls below the threshold $\text{Th}^\tau$ determined by the Planner, the Prompt Reviser is invoked for another generation. Otherwise, the generated image or video $y^\tau$, the VQA score $v^\tau$, the final evaluation score $f^\tau$, and the interpretable reasoning $r$ together form the contents $C = \{y^\tau, v^\tau, f^\tau, r\}$.

\textbf{(5) Prompt Reviser:}
The Prompt Reviser is activated when the generated result fails to meet the required threshold. It takes as input the generated content $y^\tau$, evaluation reasoning $r$ and the prompt $t_{e/r}$ used in the last generation. to identify deficiencies and rewrite the prompt using an MLLM. The revised prompt $t_r$ is then passed back to the Content Generator for another generation iteration, enabling closed-loop adaptive refinement. This process can be expressed as
\begin{equation}
    t_r = PromptReviser(y^\tau, r, t_{e/r})
\end{equation}
Such iterative refinement ensures that the generated content progressively aligns more closely with user preferences and quality expectations.

\subsection{Summerizer: Response Results and Reflection}
The Summarizer serves as the final component in the MagicWand pipeline, consolidating the outputs of the generation and evaluation processes. Based on the contents $C$ from Executor, Summarizer produces a structured summary $S(C)$ to user and collect the preference score $u^\tau$. The generated sample $y^\tau$ with its VQA score $v^\tau$ and $u^\tau$ are then used to update the memory $M$, enabling the system to refine future generations and adapt more effectively to the user preferences.

\section{UniPreferBench}
To evaluate MagicWand, we construct UniPreferBench, the first benchmark for assessing user preference alignment across diverse AIGC tasks. It enables systematic evaluation of both the generation and evaluation models, measuring how effectively they capture and satisfy user preferences consistently across different task types.

\subsection{Data Collection} 
\begin{table*}[h]
\caption{Comparison results on different AIGC methods. $\heartsuit$ open-source methods, $\spadesuit$ close-source methods. The best results are highlighted in \mredbf{red}, the second-best results are highlighted in \mbluebf{blue}, and the third-best results are highlighted in \textbf{black}.}
\centering
\belowrulesep=0pt
\aboverulesep=0pt
\setlength{\tabcolsep}{3pt} 
 \resizebox{1\textwidth}{!}{\begin{tabular}{l||cccccccccc||c}
\toprule
&\multicolumn{10}{c}{Image Generation (T2I)}\\
\cmidrule(lr){2-11}
Model/Categories & Single& Two& Multiple& Color & Light & Scene & Style & OCR & Action & Expression & Overall \\

\midrule
$\heartsuit$OmniGen2 \cite{omnigen2} & 75.68 & 72.06 & 69.72 & 77.74 & 70.30 & 71.37 & 75.70 & 70.75 & 68.81 & 69.64 & 72.18 \\
$\heartsuit$Qwen-Image \cite{qwenedit} & 78.12 & 75.45 & 71.14 & 79.21 & \textbf{75.29} & \textbf{76.60} & 74.54 & 72.08 & \textbf{74.03} & 75.24 & 75.17 \\
$\spadesuit$FLUX-Kontext \cite{fluxkontext}& 76.89 & 74.48 & 70.02 & 78.11 & 72.82 & 73.40 & 74.85 & 70.48 & 70.08 & \textbf{76.93} & 73.81 \\
$\spadesuit$SeedDream4 \cite{seedream4}& \textbf{80.64} & \textbf{77.42} & \textbf{72.26} & \textbf{81.45} & 74.17 & 74.26 & \textbf{76.82} & \textbf{71.07} & 73.53 & 76.42 & \textbf{75.81} \\
\hdashline
\noalign{\vspace{0.5pt}}
\rowcolor{gray!20}  
MagicWand (Qwen-Image) & \mbluebf{84.68} & \mbluebf{83.05} & \mbluebf{80.46} & \mbluebf{86.44} & \mbluebf{84.65} & \mbluebf{83.83} & \mbluebf{85.41} & \mredbf{81.76} & \mbluebf{80.85} & \mbluebf{83.38} & \mbluebf{83.45} \\
\rowcolor{gray!20}  
MagicWand (SeedDream4) & \mredbf{86.11} & \mredbf{84.01} & \mredbf{81.96} & \mredbf{88.38} & \mredbf{83.54} & \mredbf{82.83} & \mredbf{86.88} & \mbluebf{80.05} & \mredbf{82.72} & \mredbf{84.06} & \mredbf{84.06} \\
\midrule
&\multicolumn{10}{c}{Image Editing (I2I)}\\
\cmidrule(lr){2-11}
Model/Categories & Addition & Removal & Replacement & Color & Light & Background & Style & OCR & Action & Expression & Overall \\
\midrule
$\heartsuit$OmniGen2 \cite{omnigen2}& 75.23 & 71.20 & 74.06 & 75.78 & 70.35 & 70.30 & 72.21 & 65.37 & 67.75 & 72.03 & 71.43 \\
$\heartsuit$Qwen-Edit \cite{qwenedit}& \textbf{78.34} & 74.69 & 76.05 & 78.80 & 72.49 & 75.53 & 74.01 & 70.92 & 72.99 & 75.83 & 74.96 \\
$\spadesuit$FLUX-Kontext \cite{fluxkontext}& 76.31 & 73.21 & 74.38 & 77.11 & 74.17 & 73.63 & 72.03 & 68.86 & 70.08 & \textbf{76.97} & 73.68 \\
$\spadesuit$NanoBanana \cite{nanobanana}& 78.08 & \textbf{75.93} & \textbf{77.52} & \textbf{80.82} & \textbf{74.27} & \textbf{76.50} & 75.79 & 72.80 & 74.97 & 76.28 & \textbf{76.30} \\
\hdashline
\noalign{\vspace{0.5pt}}
\rowcolor{gray!20}  
MagicWand (Qwen-Edit) & \mredbf{82.66} & \mbluebf{81.77} & \mbluebf{82.89} & \mbluebf{86.52} & \mbluebf{80.66} & \mbluebf{82.53} & \mbluebf{81.23} & \mbluebf{79.57} & \mredbf{80.60} & \mbluebf{82.93} & \mbluebf{82.14} \\
\rowcolor{gray!20}  
MagicWand (NanoBanana) & \mbluebf{81.17} & \mredbf{83.78} & \mredbf{83.23} & \mredbf{87.25} & \mredbf{82.48} & \mredbf{84.53} & \mredbf{82.80} & \mredbf{80.88} & \mbluebf{80.45} & \mredbf{83.97} & \mredbf{83.05} \\

\midrule
&\multicolumn{10}{c}{Text-guided Video Generation (T2V)}\\
\cmidrule(lr){2-11}
Model/Categories & Single& Two& Multiple& Color & Light & Scene & Style & OCR & Action & Expression & Overall \\
\midrule
$\heartsuit$HunYuanVideo \cite{hunyuanvideo}& 68.33 & 56.90 & 52.81 & 74.19 & 68.99 & 52.82 & 58.70 & 65.15 & 71.25 & 67.77 & 63.69 \\
$\heartsuit$CogVideo X1.5 \cite{cogvideox}& 69.01 & 54.42 & 51.97 & \textbf{75.08} & 60.45 & 63.84 & 53.46 & 61.47 & 68.52 & 65.61 & 62.38 \\
$\spadesuit$Sora2 \cite{sora}& \textbf{72.31} &\textbf{ 64.51} & \textbf{65.71} & 71.12 & \textbf{72.26} & 63.63 & \textbf{67.05} & \textbf{71.52} & \textbf{76.17} & \textbf{75.82} & \textbf{70.01} \\
$\spadesuit$PixVerse \cite{pixverse}& 70.91 & 57.04 & 60.07 & 74.47 & 69.02 & \textbf{68.46} & 66.56 & 66.33 & 71.53 & 72.75 & 67.71 \\
\hdashline
\noalign{\vspace{0.5pt}}
\rowcolor{gray!20}  
MagicWand (HunYuanVideo) & \mbluebf{76.91} & \mbluebf{73.04} & \mbluebf{72.70} & \mbluebf{77.95} & \mredbf{78.89} & \mbluebf{75.99} & \mbluebf{76.23} & \mbluebf{73.18} & \mbluebf{76.10} & \mbluebf{75.66} & \mbluebf{75.66} \\
\rowcolor{gray!20}  
MagicWand (Sora2) & \mredbf{80.89} & \mredbf{75.40} & \mredbf{73.31} & \mredbf{80.93} & \mbluebf{77.45} & \mredbf{79.64} & \mredbf{81.50} & \mredbf{76.17} & \mredbf{79.02} & \mredbf{77.23} & \mredbf{78.15} \\

\midrule
&\multicolumn{10}{c}{Image-guided Video Generation (I2V)}\\
\cmidrule(lr){2-11}
Model/Categories & Single& Two& Multiple& Color & Light & Scene & Style & OCR & Action & Expression & Overall \\
\midrule
$\heartsuit$HunYuan-I2V \cite{hunyuanvideo}& 69.97 & 63.97 & 66.10 & 72.48 & 65.16 & 68.49 & \textbf{69.83} &\textbf{65.97} & 76.45 & 74.03 & 69.24 \\
$\heartsuit$CogVideo X1.5-I2V \cite{cogvideox}& 66.90 & 62.60 & 65.17 & 68.47 & 69.32 & 67.50 & 60.16 & 53.56 & 74.94 & 75.45 & 66.41 \\
$\spadesuit$Sora2 \cite{sora}& \textbf{74.81} & 73.08 & \textbf{77.68} & 77.88 & 70.37 & 70.25 & 68.39 & 64.41 & \textbf{78.87} & \textbf{75.96} & \textbf{73.17} \\
$\spadesuit$PixVerse \cite{pixverse}& 70.80 & \textbf{74.94} & 73.36 & \textbf{79.10} & \textbf{71.85} & \textbf{70.63} & 64.88 & 59.84 & 75.30 & 73.49 & 71.42 \\
\hdashline
\noalign{\vspace{0.5pt}}
\rowcolor{gray!20}  
MagicWand (HunYuan-I2V) & \mbluebf{77.76} & \mbluebf{76.27} & \mbluebf{75.59} & \mbluebf{82.88} & \mbluebf{75.76} & \mbluebf{77.25} & \mredbf{75.85} & \mbluebf{74.01} & \mbluebf{82.86} & \mbluebf{80.08} & \mbluebf{77.83} \\
\rowcolor{gray!20}  
MagicWand (Sora2) & \mredbf{82.91} & \mredbf{80.91} & \mredbf{79.73} & \mredbf{83.05} & \mredbf{78.68} & \mredbf{80.53} & \mbluebf{74.48} & \mredbf{74.85} & \mredbf{83.53} & \mredbf{81.28} & \mredbf{80.00} \\

\midrule
&\multicolumn{10}{c}{Video Editing (V2V)}\\
\cmidrule(lr){2-11}
Model/Categories & Addition & Removal & Replacement & Color & Light & Background & Style & OCR & Action & Expression & Overall \\
\midrule
$\heartsuit$InsV2V \cite{insv2v}& 52.51 & 57.10 & 54.20 & 61.11 & 46.60 & 60.76 & 53.82 & 32.52 & 55.66 & 61.09 & 53.54 \\
$\heartsuit$InsViE \cite{insvie}& 57.10 & 62.82 & 61.43 & 64.53 & 57.41 & 62.35 & 56.11 & 30.43 & 57.17 & 59.51 & 56.89 \\
$\heartsuit$Ditto \cite{ditt} & 66.81 & \textbf{69.18} & 68.17 & 72.86 & \textbf{64.91} & 62.52 & \textbf{60.24} & 48.91 & 60.69 & 65.73 & 64.00 \\
$\spadesuit$Gen4 \cite{gen4}& \textbf{67.86} & 68.72 & \textbf{71.21} & \textbf{73.96} & 63.18 & \textbf{64.61} & 58.27 & \textbf{58.05} & \textbf{65.27} & \textbf{67.77} & \textbf{65.89} \\
\hdashline
\noalign{\vspace{0.5pt}}
\rowcolor{gray!20}  
MagicWand (Ditto) & \mbluebf{75.56} & \mbluebf{74.66} & \mbluebf{74.10} & \mbluebf{76.25} & \mbluebf{74.08} & \mbluebf{75.16} & \mbluebf{72.96} & \mbluebf{65.79} & \mbluebf{71.49} & \mbluebf{72.20} & \mbluebf{73.23} \\
\rowcolor{gray!20}  
MagicWand (Gen4) & \mredbf{78.54} & \mredbf{79.84} & \mredbf{76.61} & \mredbf{77.28} & \mredbf{74.28} & \mredbf{76.96} & \mredbf{73.62} & \mredbf{68.22} & \mredbf{74.23} & \mredbf{73.26} & \mredbf{75.28} \\
\bottomrule
\end{tabular}}
\label{gen_comparison}
\end{table*}
Taking both practicality and popularity into account, we select 10 representative tasks for each generation type—T2I, I2I, T2V, I2V, and V2V. For T2I, T2V and I2V, the tasks include single class, two class, multiple class, color, light, scene, style, OCR (Optical Character Recognition), action, and expression. For I2I and V2V, the tasks cover addition, removal, replacement, color, light, background, style, OCR, action, and expression. Each category contains 20 samples. All source images and videos are collected from the Internet, with image resolutions above 512×512 and video clips lasting 5–8 seconds at 30 fps. The generation prompts are first produced by the advanced MLLM Qwen3-VL \cite{qwen3} and subsequently refined by human annotators.

Based on the source images/videos and paired prompts, we utilize our MagicWand and other open-source and closed-source generation methods to perform the generation tasks. Specifically, for T2I, we use MagicWand (Qwen-Image), MagicWand (SeedDream4), Qwen-Image~\cite{qwenedit}, SeedDream4~\cite{seedream4}, OmniGen2~\cite{omnigen2}, and FLUX-Kontext~\cite{fluxkontext}; for I2I, we select MagicWand (Qwen-Edit), MagicWand (NanoBanana), Qwen-Edit~\cite{qwenedit}, NanoBanana~\cite{nanobanana}, OmniGen2~\cite{omnigen2}, and FLUX-Kontext~\cite{fluxkontext}; for T2V, we select MagicWand (HunYuanVideo), MagicWand (Sora2), HunYuanVideo~\cite{hunyuanvideo}, Sora2~\cite{sora}, CogVideo X1.5~\cite{cogvideox}, and Pixverse~\cite{pixverse}; for I2V, we select MagicWand (HunYuan-I2V), MagicWand (Sora2), HunYuan-I2V~\cite{hunyuanvideo}, Sora2~\cite{sora}, CogVideo X1.5-I2V~\cite{cogvideox}, and PixVerse~\cite{pixverse}; for V2V, we utilize MagicWand (Ditto), MagicWand (Gen4), Ditto~\cite{ditt}, Gen4~\cite{gen4}, InsViE~\cite{insvie}, and InsV2V~\cite{insv2v}. Notably, MagicWand generates diverse contents based on individual user preferences, enabling personalized results across users. To produce AIGC with MagicWand, we recruited 20 users who first completed a few-shot preference initialization process. Afterward, we used MagicWand to generate contents conditioned on their personalized preferences. In total, we generated 40K (200 samples$\times$5 tasks$\times$2 methods$\times$20 users) images and videos using our method, along with 4K (200 samples$\times$5 tasks$\times$4 methods) additional samples produced by other generation methods.

\subsection{Subjective Experiment Setup}
\begin{table*}[t]
\caption{Comparison results on different AIGC methods. $\heartsuit$ open-source MLLMs, $\spadesuit$ close-source MLLMs, $\clubsuit$ benchmark-based evaluation score. The best results are highlighted in \mredbf{red}, and the second-best results are highlighted in \mbluebf{blue}. State-of-the-art evluation methods are selected for comparison, including GPT-4 \cite{chatgpt4o}, Gemini \cite{nanobanana}, Qwen3-VL \cite{qwen3}, CLIPScore \cite{clipscore}, LMM4LMM \cite{lmm4lmm}, EditScore \cite{editscore}, LMM4Edit \cite{lmm4edit}, VBench \cite{vbench}, LOVE \cite{love}, VBench-I2V \cite{vbench}, VideoScore \cite{videoscore}, VEBench \cite{vebench}, TDVE-assessor \cite{tdve}.}
\centering
\belowrulesep=0pt
\aboverulesep=0pt
 \resizebox{1\textwidth}{!}{
\begin{tabular}{l||ccc||l||ccc||l||ccc}
\toprule
 \multicolumn{2}{l}{T2I}&&&\multicolumn{2}{l}{I2I}&&&\multicolumn{2}{l}{T2V}\\
 
\midrule

 Model/Metric& SRCC & KRCC & PLCC &  Model/Metric & SRCC & KRCC & PLCC &  Model/Metric & SRCC & KRCC & PLCC \\
   \midrule
$\spadesuit$GPT-4 & 0.6325 & 0.5377 & 0.6575 & $\spadesuit$GPT-4 & 0.6780 & 0.6091 & 0.6908 & $\spadesuit$GPT-4 & 0.6082 & 0.5181 & 0.6081 \\
$\spadesuit$Gemini & \mbluebf{0.6616} & 0.5323 & \mbluebf{0.6798} & $\spadesuit$Gemini & 0.6716 & 0.6247 & 0.6937 & $\spadesuit$Gemini & 0.5995 & 0.5478 & 0.5918 \\
$\heartsuit$Qwen3-VL & 0.5687 & 0.5340 & 0.6240 & $\heartsuit$Qwen3-VL & 0.6847 & 0.6398 & 0.7075 & $\heartsuit$Qwen3-VL & 0.6022 & 0.5432 & 0.6226 \\
$\clubsuit$CLIPScore & 0.5956 & 0.5108 & 0.6505 & $\clubsuit$EditScore & 0.6734 & 0.6367 & 0.6937 & $\clubsuit$VBench & 0.5833 & 0.5336 & 0.5457 \\
$\clubsuit$LMM4LMM & 0.6528 & \mbluebf{0.5415} & 0.6532 & $\clubsuit$LMM4Edit & \mbluebf{0.6904} & \mbluebf{0.6422} & \mbluebf{0.7078} & $\clubsuit$LOVE & \mbluebf{0.6418} & \mbluebf{0.5744} & \mbluebf{0.6591} \\
\hdashline
\noalign{\vspace{0.5pt}}
\rowcolor{gray!20}  
MagicWand & \mredbf{0.7352} & \mredbf{0.7013} & \mredbf{0.7509} & MagicWand & \mredbf{0.7444} & \mredbf{0.7094} & \mredbf{0.7602} & MagicWand & \mredbf{0.7228} & \mredbf{0.7099} & \mredbf{0.7432} \\
\midrule
 \multicolumn{2}{l}{I2V}&&&\multicolumn{2}{l}{V2V}&&&\multicolumn{2}{l}{Overall}\\
 \midrule
  Model/Metric & SRCC & KRCC & PLCC &  Model/Metric  & SRCC & KRCC & PLCC &   Model/Metric & SRCC & KRCC & PLCC \\
  \midrule
$\spadesuit$GPT-4 & 0.6522 & 0.5538 & \mbluebf{0.6664} & $\spadesuit$GPT-4 & 0.6632 & 0.6477 & 0.6806 & $\spadesuit$GPT-4 & \mbluebf{0.6548} & \mbluebf{0.5899} & \mbluebf{0.6711} \\
$\spadesuit$Gemini & 0.6470 & 0.5237 & 0.6618 & $\spadesuit$Gemini & 0.6563 & 0.6390 & 0.6713 & $\spadesuit$Gemini & 0.6385 & 0.5637 & 0.6682 \\
$\heartsuit$Qwen3-VL & 0.5254 & 0.5099 & 0.5238 & $\heartsuit$Qwen3-VL & 0.6551 & 0.6286 & 0.6892 & $\heartsuit$Qwen3-VL & 0.6225 & 0.5225 & 0.6528 \\
$\clubsuit$VBench-I2V & 0.6133 & 0.5210 & 0.6553 & $\clubsuit$VEBench & 0.6751 & 0.6486 & 0.6989 & $\heartsuit$InternVL3.5 & 0.6090 & 0.5325 & 0.6489 \\
$\clubsuit$VideoScore & \mbluebf{0.6529} & \mbluebf{0.5941} & 0.6637 & $\clubsuit$TDVE-assessor & \mbluebf{0.7151} & \mbluebf{0.6886} & \mbluebf{0.7289} & $\heartsuit$MiniCPM-V4.5 & 0.6056 & 0.5298 & 0.6417 \\
\hdashline
\noalign{\vspace{0.5pt}}
\rowcolor{gray!20}  
MagicWand & \mredbf{0.7399} & \mredbf{0.7026} & \mredbf{0.7558} & MagicWand & \mredbf{0.7561} & \mredbf{0.7083} & \mredbf{0.7685} & MagicWand & \mredbf{0.7425} & \mredbf{0.7080} & \mredbf{0.7661} \\
\bottomrule
\end{tabular}}

\label{eval_comparison}
\end{table*}
To evaluate the generated contents, we firstly conduct a subjective quality assessment experiment using the UniPreferBench, which is designed to capture user preferences across different AIGC tasks. We recruit the same 20 participants who completed the few-shot preference initialization during the generation process. In each evaluation, participants are presented with the source image/video (if exist), the corresponding generated image/video, and the associated generation prompt. They are asked to rate the preference alignment score of each sample in the range 0-100.

The experiment is implemented using a Python-based graphical user interface (GUI) displayed on a calibrated LED monitor with a resolution of 3840$\times$2160. All images and videos are shown at a resolution of 512$\times$512 in randomized order. Participants are seated approximately 2 feet from the display in a controlled environment. To mitigate fatigue, the study is divided into 40 sessions, each lasting no more than 30 minutes. Each image/video generated by other methods is evaluated by all 20 participants, while each image/video generated by MagicWand is evaluated only by its corresponding user.

\section{Experiment}
In this section, we compare the generation and evaluation performance of our MagicWand with other state-of-the-art methods.
\subsection{Experiment Setup}
For generation, we utilize user preference scores as metric. For evaluation, to calculate the correlation between the predicted scores and the ground-truth user preference score, we use three evaluation metrics, including Spearman Rank Correlation Coefficient (SRCC), Pearson Linear Correlation Coefficient (PLCC), and Kendall’s Rank Correlation Coefficient (KRCC). All models are implemented in PyTorch and evaluated on a 40GB NVIDIA RTX A100 GPU.

\subsection{Benchmarking Generation Performance}

We compute normalized preference scores for each user and then average these scores across all users. Based on these results, we compare the generation ability of MagicWand with other state-of-the-art AIGC methods, as shown in Table~\ref{gen_comparison}. First, our MagicWand achieves the highest user preference scores across all AIGC tasks and sub-tasks. Among the baselines, SeedDream4~\cite{seedream4} obtains the best results for T2I, NanoBanana~\cite{nanobanana} performs best for I2I, and Sora2~\cite{sora} achieves the best results for both T2V and I2V. Gen4 achieves the best performance for V2V. Notably, all these methods are closed-source. However, our method combined with open-source generation models surpasses these closed-source systems, showing that users can freely utilize our approach while obtaining competitive or even superior performance.

We also find that AI-generated videos generally receive lower preference scores than AI-generated images, likely due to unnatural motion or temporal inconsistency in video generation. Moreover, editing tasks such as I2I and V2V are more challenging to satisfy users compared with generation tasks like T2I, T2V, and I2V, indicating that editing instructions introduce more difficulty for AIGC methods to generate user-preferred contents. This issue becomes more prominent in complex editing scenarios such as OCR-based editing or action modification.

\subsection{Benchmarking Evaluation Performance}
We further compare the evaluation ability of MagicWand with other advanced AIGC evaluation methods, and the results are presented in Table~\ref{eval_comparison}. We compute the SRCC, KRCC, and PLCC between the predicted scores and each user’s preference scores, and then report the averaged results. MagicWand consistently achieves the best performance across all AIGC tasks and even surpasses GPT-4o~\cite{chatgpt4o}, demonstrating its strong alignment with user preferences. In addition, we observe that benchmark-based evaluation methods often outperform advanced MLLMs. This is largely because benchmark-driven approaches rely on curated, task-specific reference data and human-annotated ground truths, which offer more stable assessments. However, since most benchmarks reflect average human perception, their ability to align with individual user preferences remains limited, highlighting the need for future work on personalized evaluation modeling.

\section{Conclusion}
In this work, we introduce MagicWand, a multi-task agent for generating and evaluating AI-generated content in alignment with user preferences. We also present UniPreferBench, a large-scale benchmark containing 120K user preference scores across various AIGC tasks, validating that MagicWand excels in both generation and evaluation, achieving state-of-the-art performance.
\bibliography{example_paper}
\bibliographystyle{icml2025}
\end{document}